\documentclass[10pt,twocolumn,letterpaper]{article}

\usepackage{iccv}
\usepackage{times}
\usepackage{epsfig}
\usepackage{graphicx}
\usepackage{amsmath}
\usepackage{amssymb}

\usepackage{times}
\usepackage{epsfig}
\usepackage{graphicx}
\usepackage{amsmath}
\usepackage{amssymb}
\usepackage{algorithm}
\usepackage{algpseudocode}
\usepackage{threeparttable}
\usepackage{multirow}
\usepackage{subfigure}
\usepackage{booktabs}
\usepackage{caption}
\usepackage{mathrsfs}
\usepackage{bbding}
\usepackage{authblk}
\usepackage[breaklinks=true,bookmarks=false]{hyperref}

\iccvfinalcopy % *** Uncomment this line for the final submission

\def\iccvPaperID{3076} % *** Enter the ICCV Paper ID here

\ificcvfinal\fi

\begin{document}

%%%%%%%%% TITLE
\title{Adaptive Hierarchical Graph Reasoning with Semantic Coherence for Video-and-Language Inference}

	\author{
	\centerline{Juncheng Li$^1$\quad Siliang Tang$^1$\thanks{Siliang Tang is the corresponding author.}\quad Linchao Zhu$^2$ \quad Haochen Shi$^3$\quad Xuanwen Huang$^1$}
	\centerline{Fei Wu$^1$\quad Yi Yang$^1$\quad Yueting Zhuang$^1$\quad }
	\centerline{$^1$Zhejiang University\quad $^2$ReLER, University of Technology Sydney\quad $^3$Universit\'{e} de Montr\'{e}al}
	\centerline{{\tt\small {\{junchengli, siliang, xwhuang, wufei, yangyics, yzhuang\}}@zju.edu.cn , {linchao.zhu}@uts.edu.au}}
	\centerline{{\tt\small haochen.shi@umontreal.ca}}
	}

\maketitle

\ificcvfinal\thispagestyle{empty}\fi

%%%%%%%%% ABSTRACT
	\begin{abstract}
	Video-and-Language Inference is a recently proposed task for joint video-and-language understanding. This new task requires a model to draw inference on whether a natural language statement entails or contradicts a given video clip. In this paper, we study how to address three critical challenges for this task: judging the global correctness of the statement involved multiple semantic meanings, joint reasoning over video and subtitles, and modeling long-range relationships and complex social interactions. First, we propose an adaptive hierarchical graph network that achieves in-depth understanding of the video over complex interactions. Specifically, it performs joint reasoning over video and subtitles in three hierarchies, where the graph structure is adaptively adjusted according to the semantic structures of the statement. Secondly, we introduce semantic coherence learning to explicitly encourage the semantic coherence of the adaptive hierarchical graph network from three hierarchies. The semantic coherence learning can further improve the alignment between vision and linguistics, and the coherence across a sequence of video segments. Experimental results show that our method significantly outperforms the baseline by a large margin.
\end{abstract}

%%%%%%%%% BODY TEXT
\vspace{-0.5cm}
\section{Introduction}
Understanding video story involves analyzing and simulating human vision, language, thinking, and behavior, which is a significant challenge to current machine learning technology~\cite{heo2019constructing}. Recently, with the advances of large-scale video datasets~\cite{abu2016youtube, caba2015activitynet, cheng2014temporal, kay2017kinetics, wang2016walk}, joint video-and-language understanding has received increased attention. Several video-and-language tasks have been proposed, such as video captioning~\cite{guadarrama2013youtube2text, venugopalan2015sequence, zhang2020relational, gan2017semantic, krishna2017dense, gan2017stylenet, pu2016adaptive, wang2019vatex}, text-to-video temporal grounding~\cite{gao2017tall, anne2017localizing, chen2018temporally, lei2020tvr, mun2020local, zhang2021consensus, zhu2020actbert}, and video question answering~\cite{lei2018tvqa, zhang2019frame, tapaswi2016movieqa, jang2017tgif, kim2017deepstory, mun2017marioqa}. In particular, Video-and-Language Inference~(VLI)~\cite{liu2020violin} is a recently proposed task to foster deeper investigations in video-and-language understanding. Given a video clip with aligned subtitles and a natural language statement based on the video content, a model needs to infer whether the statement entails or contradicts the given video clip. To support the study of this new task, a large-scale dataset, named VIOLIN (VIdeO-and-Language INference), is introduced.

%VLI requires not only explicit information in the video but also more sophisticated reasoning skills, such as inferring reasons and interpreting human emotions. 

\begin{figure}[!t]
	\centering
	\includegraphics[width=\linewidth]{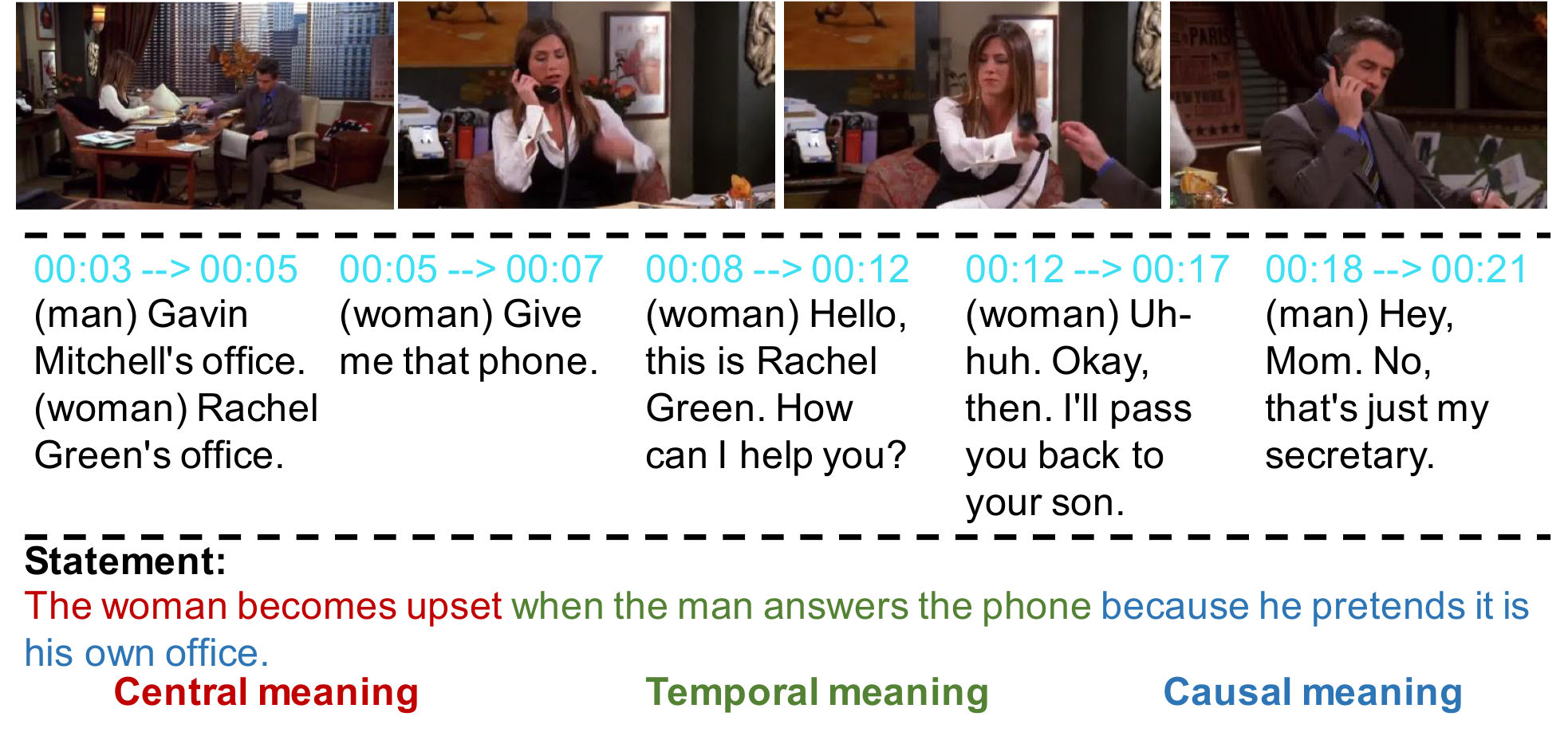}
	\vspace{-0.6cm}
	\caption{The first two rows show a video clip paired with its aligned subtitles. The third row shows a statement with multiple semantic meanings.}
	\label{demo}
	\vspace{-0.7cm}
\end{figure}

Compared with TVQA/video captioning where most QA pairs/captions focus on identifying explicit visual cues~(\eg, objects, actions, persons), VLI is more challenging and requires more sophisticated reasoning skills, such as interpreting human emotions and relations, understanding the events, and inferring causal relations of events throughout the video. First, a single statement may involve multiple semantic meanings, making it harder to judge the global correctness. As demonstrated in Figure \ref{demo}, the statement consists of three semantic phrases. If the model recognizes the central meaning and the temporal meaning but ignores the causal meaning, it may make a wrong prediction. Secondly, VLI requires jointly reasoning over video and subtitles to achieve in-depth understanding of complex plots. To infer the causal meaning that the man pretends it is his own office, the model needs to jointly understand information from the video part and subtitle part. From the video part, the man and the woman are in the same office, and the man takes the phone from the woman. From the subtitle part, the man lies that the woman is his secretary. Only by combining the context from both the video and subtitle can the model further draw the inference. Thirdly, VLI requires reasoning for diverse interactions among characters and complex event dynamics over diverse scenarios. The VIOLIN dataset is collected from diverse sources to cover realistic visual scenes, including 5885 movie clips in addition to TV shows used in TVQA. The average clip length is 35.20s,  while the length of most clips in TVQA is less than 15s.

%(e.g., interpreting human emotions and relations, understanding the events, and inferring causal relations of events throughout the video). 
%Secondly, videos are collected from TV shows and movies, which contain complex social interactions and diverse scenarios. This requires a model to reason over the relationship among multiple visual scenes to draw a global video understanding.

%We convert video frames and subtitle words in each modality into a graph, where the subtitle words in the same sentence are aligned with a sequence of visual frames whose timestamps overlap with the subtitle timestamp. Each frame-sub

In this paper, we propose a novel adaptive hierarchical graph reasoning with semantic coherence approach to overcome the aforementioned challenges. First, we introduce an adaptive graph construction mechanism to identify the multiple semantic meanings of the statement. This enables our approach to adaptively adjust the graph structure according to the semantic structures of the statement for the global correctness. Then, we present an adaptive hierarchical graph network~(AHGN) to jointly reason over video and subtitles and model the complex social interactions. Specifically, we perform adaptive graph reasoning in three hierarchies: 1) segment-level reasoning, which achieves in-depth understanding of the video segments via utilizing the inherent alignment and complementary nature between visual frames and subtitles; 2) temporal-level reasoning, which models the long-range dependencies and diverse interactions between different segments to draw a global video understanding;  3) global-level reasoning, which judges the global correctness of the statement by incorporating the inferences from different reasoning steps.

Furthermore, the semantic coherence throughout AHGN is crucial to achieving global understanding of the video. Therefore, we introduce a novel semantic coherence learning~(SCL) method to encourage the cross-modal semantic coherence at the segment level and the cross-level semantic coherence between temporal level and global level. Specifically, the semantic coherence learning contains two regularization terms: an optimal transport distance term that measures the cross-modality alignment between the visual nodes and subtitle nodes, and a mutual information term that evaluates the semantic coherence between the temporal nodes and global nodes.

The experiments show that our approach significantly outperforms the baselines by a large margin, and further ablation study demonstrates the effectiveness of each component. In summary, our contributions are mainly three folds:

%Our contributions can be summarized as follows:

\begin{itemize}
	\item We propose a novel adaptive hierarchical graph network~(AHGN) that performs joint reasoning over video and subtitles in three hierarchies, where the graph reasoning structure is adaptively adjusted according to the semantic structures of the statement.
	
	%thus achieving in-depth story-level semantics understanding.
	
	\item Our semantic coherence learning~(SCL) method improves the alignment between video and subtitles, and the coherence across a sequence of video segments.
	
	%\item We introduce a dynamic adaptive semantic inference module that adaptively extracts a variable number of semantic queries from a statement for the following hierarchical graph reasoning.
	
	%\item We propose a novel hierarchical graph reasoning network that jointly reasons over video and subtitles in three hierarchies under the guidance of the semantic queries, of which we introduce the semantic coherence learning to encourage the semantic coherence among the three hierarchies.
	
	\item Extensive experiments show that our method significantly outperforms the baseline by a large margin.
	
	%	\item We conduct extensive experiments to evaluate our approach and show that our method significantly outperforms the baseline by a large margin.
	
\end{itemize}

\begin{figure*}
	\begin{center}
		\includegraphics[width=\textwidth]{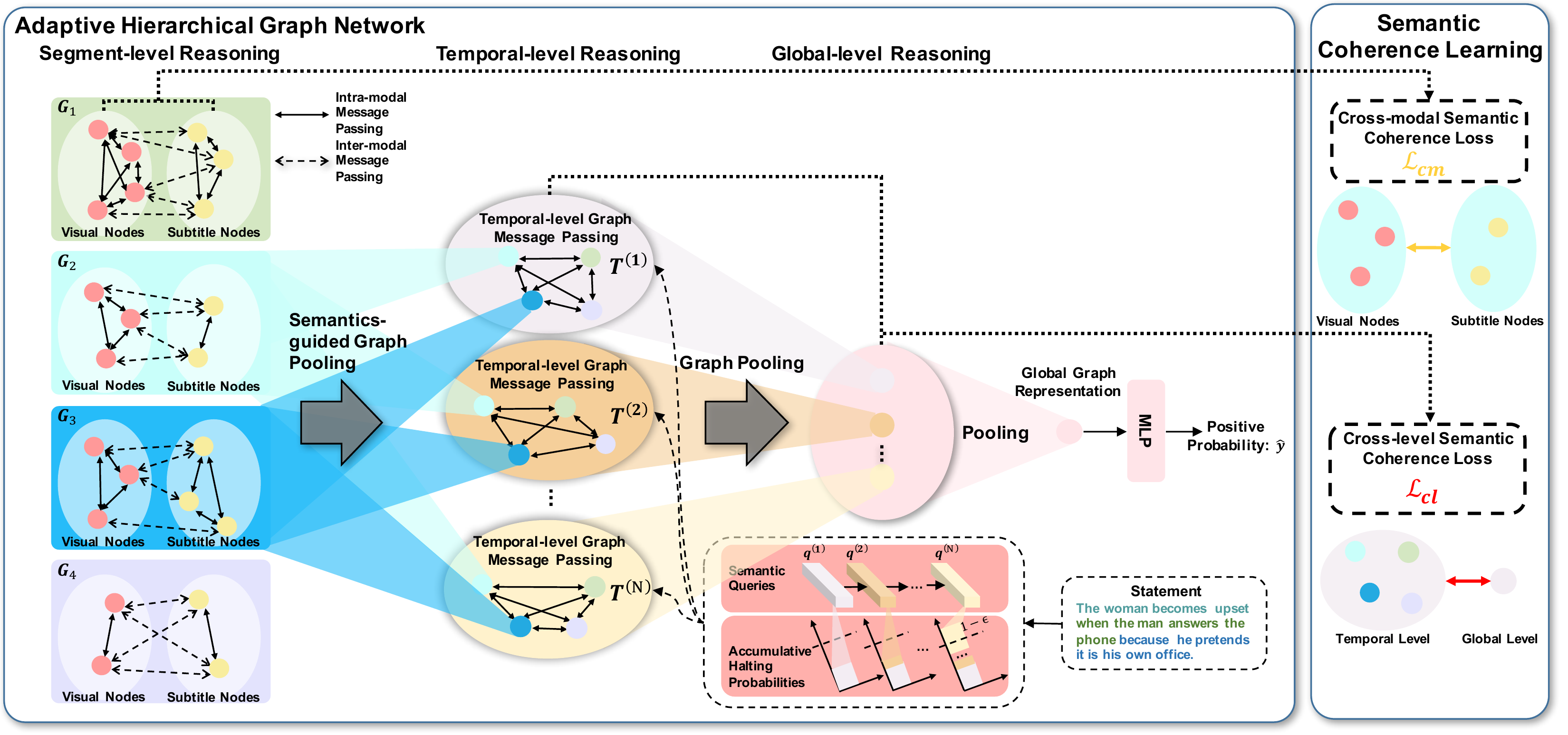}
	\end{center}
	\vspace{-0.5cm}
	\caption{Overview of the proposed framework. Each sub-graph $G_i$ is constructed with subtitle word nodes $\{s_i\}$ and time-aligned visual frame nodes $\{v_i\}$. Our AHGN performs joint reasoning over video and subtitles in three hierarchies. $T^{(i)}$ represents a temporal-level sub-graph, which is adaptively constructed according to the semantic structures of the statement. The semantic coherence learning explicitly promotes the cross-modal and cross-level semantic coherence of AHGN.}
	\label{overview}
	\vspace{-0.5cm}
\end{figure*}

%Given a statement and a video paired with the subtitles, the adaptive hierarchical graph network adaptively performs joint reasoning over video and subtitles in three hierarchies according to the semantic structure of the statement. The semantic coherence learning provides explicit training signals to promote the cross-modal semantic coherence at segment level and the cross-level semantic coherence between temporal level and global level. Finally, the MLP makes the prediction based on the pooled global graph representation.

\vspace{-0.3cm}
\section{Related Work}\label{s2}
\noindent
\textbf{Visual Entailment}
Given a natural image premise and a natural language hypothesis, the goal of visual entailment~(VE)~\cite{xie2019visual} is to predict whether the image semantically entails the text. To realize this task, the SNLI-VE dataset is built based on the  Stanford Natural Language Inference corpus and Flickr30k dataset~\cite{young2014image}. Also, Suhr \textsl{et al.}~\cite{suhr2018corpus} propose a similar task to determine whether a natural language caption is true about a photograph. In contrast to the visual entailment, which is limited to a static image, video-and-language inference involves complex temporal dynamics and requires the model to understand the relationship between different visual scenes to draw the inference. In this paper, we propose an approach to model the complex inter- and intra- modality interactions and further infer in-depth rationale in three hierarchies.

\noindent
\textbf{Video-and-Language Research}\quad  Recent years have witnessed the flourishing development in vision-and-language research~\cite{antol2015vqa, gao2017tall, zhang2020devlbert, li2020unsupervised, li2019walking, guo2021semi-supervised}. Several large-scale video datasets~\cite{abu2016youtube, caba2015activitynet, cheng2014temporal, kay2017kinetics, wang2016walk} and video-and-language tasks have been proposed, such as video captioning~\cite{guadarrama2013youtube2text, ijcai2020-131, zhang2020poet, zhang2020comprehensive, venugopalan2015sequence, xu2016msr, krishna2017dense, gan2017stylenet, pu2016adaptive, wang2019vatex}, text-based video moment retrieval~\cite{gao2017tall, anne2017localizing, chen2018temporally, lei2020tvr, mun2020local}, and video question answering~\cite{lei2018tvqa, zhu2017uncovering, tapaswi2016movieqa, kai2021ask, kim2017deepstory, mun2017marioqa}. Video caption is the task of generating text descriptions from video input, text-based video moment retrieval requires localizing video segments from natural language queries, and video question answering is aimed to predict answers to natural language questions given a video as context. These tasks mainly focus on explicit factual descriptions or explicit information of the video, which hardly incorporate story-level understanding. In contrast, video-and-language inference~\cite{liu2020violin} requires not only explicit visual cues but also more sophisticated reasoning skills, such as inferring reasons and interpreting human emotions. These abilities can be used to detect anomalous intent from surveillance and discriminatory or antisocial contents from online videos, which are usually expressed implicitly. Similar to \cite{lei2018tvqa} for TVQA, the baseline model for VLI~\cite{liu2020violin} utilizes multi-stream neural network~\cite{liu2020violin, lei2018tvqa} with bidirectional attention~\cite{lei2018tvqa, seo2016bidirectional, yu2018qanet} to interact the statement with subtitles and visual frames separately, and then fuse the independent attention representations to draw the final inference. However, in our view, there are two main limitations of this type of approaches: 1) it fails to leverage the temporal alignment and complementary nature between visual frames and subtitles, which is a crucial step to achieve in-depth understanding of the video; and 2) it equally uses each word in the statement to attend every visual frame and subtitle word, and then do single-step classification, without considering explicit semantic structures.  Our work instead models the semantic alignment between visual frames and subtitles, and the graph structure is adaptively adjusted according to the semantics of the statement.

% Our work instead performs adaptive hierarchical graph reasoning over video and subtitles, which 
%
%
%introduces an adaptive hierarchical graph reasoning model to perform joint reasoning over video and subtitles, where 

%under the guidance of a variable number of semantic queries extracted from the statements.

%\vspace{-0.4cm}
\section{Method}
%Given a video clip $V$ consisting of a sequence of video frames $\{v_i\}_{i=1}^{N_v}$, paired with its aligned subtitle sentences $S=\{{s_i}, t_i^{0}, t_i^{1}\}_{i=1}^{N_s}$, and a natural language statement $H$, a model is expected to infer whether the statement is entailed or contradicted by the given video clip. The video-and-language inference task requires to joint understanding of both video and subtitles and in-depth comprehension of the interactions between visual scenes at multiple levels.

As described in Figure \ref{overview}, the adaptive hierarchical graph reasoning framework mainly consists of two components: 1)~adaptive hierarchical graph network~(Sec~\ref{s3.1}), and 2)~semantic coherence learning~(Sec~\ref{s3.2}). Given a video clip, paired with its aligned subtitles, and a natural language statement, the adaptive hierarchical graph network~(AHGN) performs reasoning in three hierarchies, of which the graph structure is adaptively adjusted according to the semantics of the statement. Finally, the prediction layer performs classification using the global graph representation. The semantic coherence learning~(SCL) is further introduced to explicitly promote the cross-modal and cross-level semantic coherence of the adaptive hierarchical graph network.

%first adaptively extracts a variable number of semantic queries according to the semantic structures of the statement. Then, the AHGN performs reasoning in three hierarchies under the guidance of the semantic queries, which enables the graph structure to be adaptive. Finally, the prediction layer performs classification using the global graph representation. The semantic coherence learning~(SCL) is further introduced to explicitly promote the cross-modal and cross-level semantic coherence of the adaptive hierarchical graph network.

%To encourage the cross-modal alignment and promote cross-level semantic coherence, we introduce the semantic coherence learning, which contains two regularization terms: an optimal transport distance term that measures the cross-modality alignment between the visual nodes and subtitle nodes, and a mutual information term that evaluates the semantic coherence between the temporal nodes and global nodes.

\subsection{Adaptive Hierarchical Graph Network}\label{s3.1}

%\vspace{-0.2cm}

\vspace{-0.2cm}
\subsubsection{Graph Construction}
%Given a pair of video clip and its associated subtitles, we first extract a sequence of visual features $\{v_i\}_{i=1}^{N_v}$ and tokenize each subtitle sentence $s_i$ into a sequence of word features $\{w_{s_i}^j\}_{j=1}^{N_{s_i}}$. We then aligh each subtitle sentence with a sequence of visual frames $V_{s_i}=\{v_{s_i}^j\}_{j=1}^{N_{v_{s_i}}}$ whose timestamps overlap with the subtitle timestamp. 

For the given video clip, we extract the visual features using ResNet101~\cite{he2016deep} trained on ImageNet~\cite{deng2009imagenet} and apply a single-layer MLP to obtain the visual node representations $V = \{v_i\}_{i=1}^{l_v}$. For its associated subtitles, we tokenize them into a word sequence and employ a pre-trained BERT~\cite{devlin2018bert} encoder, followed by a single-layer MLP, to obtain the subtitle word node representations $S =  \{s_i\}_{i=1}^{l_s}$. Each node $v_i$ corresponds to a visual frame, and each node $s_i$ corresponds to a subtitle word. We then align each subtitle sentence $S_i=\{s_i\}_{i=sst_i}^{sst_i+L}$ with a sequence of visual frames $V_i=\{v_i\}_{i=vst_i}^{vst_i+K}$ whose timestamps overlap with the subtitle timestamp ($sst_i$ is the start index of subtitle nodes in $S_i$, $L$ is the number of subtitle nodes in $S_i$, similar denotation for $V_i$). The frame-subtitle pair $<S_i, V_i>$ corresponds to a semantic segment of the video and makes up a cross-modal sub-graph $G_i$, where we perform segment-level reasoning. Thus, we split the original video-level graph into a sequence of sub-graphs~(\ie $G =\{ {G_i} \}_{i=1}^M$).

\vspace{-0.2cm}
\subsubsection{Segment-level Reasoning}
To utilize the inherent alignment and complementary nature between visual frames and subtitles,  for each segment-level sub-graph, our adaptive hierarchical graph network first models the inter- and intra- modality interactions, using the gated inter-modal message passing~(GER) and the gated intra-modal message passing~(GRA). Then, we summarize the semantic information of each sub-graph into a temporal-level node representation by semantics-guided graph pooling under the guidance of the semantic query.

%Most of the previous perform attention over the visual frames and subtitle words separately, ignoring the temporal alignment and complementary nature between visual frames and subtitles, which is a crucial step to achieve in-depth understanding of the video. Differently, for each segment-level sub-graph, our hierarchical graph network first models the inter- and intra- modality interactions, using the gated inter-modal message passing and the gated intra-modal message passing. Then, we summarize the sub-graph semantic information into a temporal-level node representation by semantics-guided graph pooling under the guidance of the semantic query.

\noindent
\textbf{Gated Inter-modal Message Passing}\quad Here we take the message passing from subtitle nodes to visual nodes as an illustration. Given segment-level sub-graph $G_i = <S_i, V_i>$, we first learn the cross-modal adjacency correlation matrix $A \in \mathbb{R}^{K \times L}$ by calculating the similarity of each pair of visual node $v_i$ and subtitle node $s_j$ as:
\begin{equation}
A = V_i^T \cdot S_i, \quad a_{ij} = v_i^T \cdot s_j
\end{equation}

\noindent
Then we compute the message $m_i^v$ that denotes the message from subtitle nodes to visual node $v_i$: $m_i^v = \sum_{j}^{L} a_{ij} \cdot s_j$. Next, we get the visual guidance $g_v$ and subtitle guidance $g_s$ by performing average pooling on the visual nodes and subtitle nodes respectively. The context gate $c_i^v$ is further determined as:

\begin{equation}
c_i^v = \sigma(W_1[g_v, v_i, g_s] + b_1)
\end{equation}

\noindent
where $W_1 \in \mathbb{R}^{d \times 3d}$, $b_1 \in \mathbb{R}^{d \times 1}$, and $\sigma(\cdot)$ denotes the sigmoid function. The context gate $c_i^v$ controls the flow of linguistic information from subtitles to vision:

\begin{equation}
\tilde{v_i}= (1-c_i^v)  \odot v_i + c_i^v \odot m_i^v
\end{equation}

\noindent
where $\odot$ denotes the Hadamard product. As a consequence, $\tilde{v_i}\in \mathbb{R}^{d \times 1}$ represents the linguistic-refined visual node, and we can obtain the visual-refined subtitle node $\tilde{s_i} \in \mathbb{R}^{d \times 1}$ in a similar manner but reversed order, and $\tilde{G_i} = <\tilde{S_i}, \tilde{V_i}>$ represents the sub-graph after GER.

\noindent
\textbf{Gated Intra-modal Message Passing}\quad For $\tilde{V_i}$ and $\tilde{S_i}$, we further refine them with intra-modal local context information using GRA. GRA is similar to the GER but models the intra-modal relations. 

The GRA first computes the intra-modal adjacency correlation matrixes $A^v \in \mathbb{R}^{K \times K}$ and $A^s \in \mathbb{R}^{L \times L}$ by calculating node-wise similarity, and then get the aggregated messages $n^v_i$ and $n^s_i$ according to the weight matrixes. Next, the GRA computes the context gates for visual/subtitle nodes based on the visual/subtitle guidance and corresponding visual/subtitle node representations respectively. Finally, GRA updates the visual nodes and subtitle nodes respectively, controlled by the context gates. As a consequence, $\hat{V_i}$ and $\hat{S_i}$ represent refined nodes of $G_i$ after inter- and intra- modality reasoning.

%then aggregate the messages $n^v_i$ and $n^s_i$ as follows:
%
%
%\begin{equation}
%	A^v = V_i^T \cdot V_i, \quad A^s = S_i^T \cdot S_i
%\end{equation}
%
%\begin{equation}
%	n_i^v = \sum_{j}^{K} a_{ij}^v \cdot v_j, \quad n_i^s = \sum_{j}^{L} a_{ij}^s \cdot s_j
%\end{equation}

%\noindent
%Next, the GRA computes the context gate $\alpha^v_i$ and $\alpha^s_i$,  and finally updates the visual nodes and subtitle nodes respectively, using the incoming messages.
%
%\begin{equation}
%	\alpha^v_i = \sigma(W_2[g_v, v_i] + b_2), \quad \alpha^s_i = \sigma(W_3[g_s, s_i] + b_3)
%\end{equation}
%
%\begin{equation}
%	\hat{v_i} = (1-\alpha^v_i)  \odot \tilde{v_i} + \alpha^v_i \odot n_i^v, \quad
%	\hat{s_i} = (1-\alpha^s_i)  \odot \tilde{s_i} + \alpha^s_i \odot n_i^s
%\end{equation}
%
%\noindent
%where $W_2, W_3 \in \mathbb{R}^{d \times 2d}$ and $b_2, b_3 \in \mathbb{R}^{d \times 1}$ are trainable parameters and $\sigma(\cdot)$ denotes the sigmoid function. As a consequence, $\hat{V_i}$ and $\hat{S_i}$ represent refined nodes of $G_i$ after inter- and intra- modality reasoning.

\noindent
\textbf{Semantics-guided Graph Pooling}\quad After obtaining segment-level refined node representations $\hat{V_i} \in \mathbb{R}^{d \times K}$ and $\hat{S_i} \in \mathbb{R}^{d \times L}$ using inter- and intra- modality local context, we further aggregate $G_i = <\hat{V_i}, \hat{S_i}>$ to a temporal-level node representation by semantics-guided graph pooling. We first extract a semantic query $q^{(n)} \in \mathbb{R}^{d \times 1}$ from the statement using attentive aggregation. Then, we employ the semantic query to attend each visual node and subtitle node independently, and then use the attention weights to aggregate $\hat{V_i}$ and $\hat{S_i}$:

\begin{equation}
C^{(n)}_i = \mathop{softmax} ({\hat{V_i}}^T(W_2q^{(n)})),\quad v^{(n)}_i  = \hat{V_i}C^{(n)}_i
\end{equation}

%\begin{equation}
%v^{(n)}_i  = \hat{V_i}C^{(n)}_i
%\end{equation}

\noindent
And we can obtain $s^{(n)}_i \in \mathbb{R}^{d \times 1}$ in a similar manner. Then, we compute the context gate $\gamma^{(n)}_i$ based on the visual guidance, subtitle guidance, and the semantic query, which controls the fusion of visual and linguistic information:

\begin{equation}
\gamma^{(n)}_i = \sigma(W_3[g_v, q^{(n)}, g_s] + b_3)
\end{equation}

\begin{equation}
t^{(n)}_i = (1 - \gamma^{(n)}_i)  \odot v^{(n)}_i + \gamma^{(n)}_i \odot s^{(n)}_i
\end{equation}

\noindent
where $W_3 \in \mathbb{R}^{d \times 3d}$ and $b_3 \in \mathbb{R}^{d \times 1}$. As a consequence, we pool each segment-level sub-graph $G_i$ to a temporal-level node $t^{(n)}_i \in \mathbb{R}^{d \times 1}$ and obtain temporal-level sub-graph $T^{(n)} \in \mathbb{R}^{d \times M}$, guided by the semantic query $q^{(n)}$. 

\vspace{-0.3cm}
\subsubsection{Temporal-level Reasoning}
The video clips are collected from TV shows and movies, which contain complex event dynamics and diverse character interactions across multiple segments. Therefore, segment-level reasoning is not sufficient, and we present temporal-level reasoning to model long-range relationships among multiple video segments to draw a global understanding. In this section, we first introduce how to construct multiple temporal-level sub-graphs adaptively, and then present how to perform reasoning on them.

\noindent
\textbf{Adaptive Temporal-level Sub-graph Construction}
Different statements may have semantic structures of varying complexity. For more complex statements, it is reasonable to construct more temporal-level sub-graphs, which focus on different semantic parts of the statement. Thus, we introduce the adaptive temporal-level sub-graph construction to adaptively adjust the  number of temporal-level sub-graphs. Concretely, we extract a variable number of semantic queries from the statement, and use them to perform semantics-guided graph pooling respectively to construct multiple temporal-level sub-graphs. The semantic queries provide guidance to their corresponding temporal-level sub-graphs about which semantic parts they should focus on.

%corresponds to the semantic part which the temporal-level sub-graph focuses on.

Given $l_h$-word statement $H = \{ h_i \}_{i=1}^{l_h}$, we extracts $N$ semantic queries $\{q^{(n)}\}_{n=1}^N$. At each step $n$, we first compute the attention weights $R^{(n)} \in \mathbb{R}^{l_h \times 1} $ based on the previous semantic query $q^{(n-1)}$ and the sentence-level embedding of the statement $g_h \in \mathbb{R}^{d \times 1}$  obtained by performing average pooling on the statement words $H$, given by:

\begin{equation}
R^{(n)} = softmax(H^T(W_r[g_h, q^{(n-1)}]))
\end{equation}

\noindent
%where $W_r  \in \mathbb{R}^{d \times 2d}$ is trainable parameters.
Then, we obtain the semantic query $q^{(n)}$ using the attention weights to summarize the statement words: $q^{(n)} = HR^{(n)}$.

To adaptively determine how many temporal-level sub-graphs are supposed to be constructed, we introduce a self-halting mechanism that outputs the probability of stopping generating more queries: $h^{(n)} = \sigma(W_hq^{(n)} + b_h)$ where $W_h \in \mathbb{R}^{d \times d}$, $b_h \in \mathbb{R}^{d \times 1}$, and the accumulative halting probability is further determined as: $P^{(n)} = \sum_{i=1}^n h^{(i)}$. When the accumulative halting probability exceeds a threshold $1 - \epsilon$ or $n$ reaches the pre-defined maximum value $N_{max}$, the process will stop. 
To promote the generation efficiency, we define a loss term based on the final number of queries: 

\begin{equation}
\mathcal{L}_{qe} =  \tau N
\end{equation}

\noindent
where $\tau$ is the query efficiency hyper-parameter.

The message passing at temporal-level graph is similar to the GRA. Given $T^{(n)}$, we first compute the adjacency correlation matrix $E^{(n)} \in \mathbb{R}^{M \times M} $, then sum up the incoming messages, and finally update the node representations using gate mechanism. After obtaining refined $\tilde{T}^{(n)}$, we further use semantic query $q^{(n)}$ to pool $\tilde{T}^{(n)}$ to a global semantic representation $o_n$:

%The whole process is summarized as follows:
%
%\begin{equation}
%	E^{(n)} = (T^{(n)})^T \cdot T^{(n)}, \quad w^{(n)}_i = \sum_{j}^{M} e_{ij}^{(n)} \cdot t_j
%\end{equation}
%
%\begin{equation}
%	\beta^{(n)}_i = \sigma(W_6[g_t^{(n)}, t_i] + b_6), \quad \tilde{t}_i = (1-\beta^{(n)}_i )  \odot t_i + \beta^{(n)}_i  \odot w^{(n)}_i
%\end{equation}
%
%\noindent
%where $W_6 \in \mathbb{R}^{d \times 2d}$ and $b_6 \in \mathbb{R}^{d \times 1}$ are trainable parameters, and $g_t^{(n)}$ is the temporal guidence from average pooling the $T^{(n)}$. After obtaining refined $\tilde{T}^{(n)}$, we further pool $\tilde{T}^{(n)}$ to a global semantic representation $o_n$ in a similar manner of semantics-guided graph pooling:

\begin{equation}
\resizebox{.88\hsize}{!}
	{
$U^{(n)}\!=\!softmax((\tilde{T}^{(n)})^T(W_4q^{(n)})),\quad o_n\!= \!\tilde{T}^{(n)}U^{(n)}$
}
\end{equation}

%\begin{equation}
%o_n  = \tilde{T}^{(n)}U^{(n)}
%\end{equation}

\noindent
where $W_4 \in \mathbb{R}^{d \times d}$ , $U^{(n)} \in \mathbb{R}^{K \times 1}$ is the attention weight vector, and $o_n \in \mathbb{R}^{d \times 1}$ is the global node representation based on semantic query $q^{(n)}$.

\vspace{-0.3cm}
\subsubsection{Global-level Reasoning}
After temporal-level reasoning, we obtain a set of $N$ global semantic representations $\{{o_i}\}_{i=1}^N$. We perform average pooling on the set of global semantic representations to generate a $d$-dimensional global graph representation capturing whole semantics of video and statement. Finally, the global graph representation is passed through the MLP with a sigmoid activation to predict the probability of the input statement being positive.

\subsection{Semantic Coherence Learning}\label{s3.2}
We specify the semantic coherence learning of AHGN as the cross-modal semantic coherence among segment-level nodes and the cross-level semantic coherence across temporal-level nodes and global-level nodes. 

%and cross-level semantic coherence, and  introduce the SCL to further promote the cross-modal semantic coherence among segment-level nodes and the cross-level semantic coherence across temporal-level nodes and global-level nodes.

\vspace{-0.3cm}
\subsubsection{Cross-modal Semantic Coherence}
To achieve in-depth understanding of the semantics in video and subtitles, a model needs to align the inter-modality semantics between visual nodes and subtitle nodes. However, most of the previous methods seek advanced attention mechanisms to simulate soft alignment, with no training signals to explicitly encourage alignment. Differently, we leverage recent advances in Optimal Transport~(OT)~\cite{chen2020graph} to encourage the cross-modal semantic coherence of each sub-graph, which can further refine the segment-level reasoning on every sub-graph $G_i$.

Optimal transport evaluates the correspondence between two distributions. OT-based learning aims to optimize distribution matching via minimizing the cost of transporting one distribution to another, which provides explicit signals to minimizing the embedding distance between the modalities. Recently, it has been explored in some fields. Liu \textsl{et al.}~\cite{liu2020semantic} model semantic correspondence as an optimal transport problem. Su \textsl{et al.}~\cite{su2015optimal} apply optimal transport to 3D shape matching and comparison. Chen \textsl{et al.}~\cite{chen2020graph} solve cross-domain alignment by minimizing the optimal transport plan between domains. Here, we adopt OT to refine the segment-level reasoning on sub-graph $G_i$. By optimizing the node distance and edge distance between visual nodes and subtitle nodes, we further foster the semantic coherence learning of gated inter- and intra- modal message passing.

%Inspired by \cite{chen2020graph}, we adapt optimal transport~(OT) to explicitly encourage semantic alignment between visual nodes and subtitle nodes at segment-level after being refined by inter- and intra-modal message passing. OT-based learning aims to optimize distribution matching via minimizing the cost of transporting one distribution to another. Here, we explicitly optimize the embedding distance between visual nodes and subtitle nodes through minimizing the cost of the learned transport plan. 

Specifically, we adapt Wasserstein distance~(WD)~\cite{luise2018differential} for node matching and Gromov-Wasserstein distance~(GWD)~\cite{peyre2016gromov, chowdhury2019gromov} for edge matching. We define two distributions $\mu_s \in P(S), \mu_v \in P(V)$ as: $\mu_s = \sum_{i=1}^n p_i^s\delta_{s_i}$ and $\mu_v = \sum_{j=1}^m p_j^v\delta_{v_i}$ where $\delta_{s_i}$ denotes the Dirac function centered on $s_i$. Without ambiguity, we reuse $m$ and $n$ to represent the number of visual nodes and subtitle nodes for simplicity. $\Pi(\mu_s, \mu_v)$ denotes all the joint distributions, with marginals $\mu_s(s)$ and $\mu_v(v)$. Let $p^s=\{p_i^s\}_{i=1}^n \in \Delta_n$ and $p^v=\{p_j^v\}_{j=1}^m \in \Delta_m$ denote the $n-$ and $m-$ dimensional weight vectors respectively, where $\sum_{i=1}^n p_i^s = \sum_{j=1}^m p_j^v = 1$, and both $p^s$ and $p^v$ are probability distributions. $\Pi(p^s, p^v) = \{T\in \mathbb{R}^{n \times m}| T1_m=p^s, T^T1_n=p^v \}$, where $T$ denotes the transport plan and $T_{ij}$ represents the amount of mass shifted from $p_i^s$ to $p_j^v$. Formally, the optimal transport distance is defined as:

\begin{small}
\begin{align}
\mathcal{D}(\mu_s&, \mu_v)\!=\!\mathop{inf}\limits_{\gamma \in \Pi(\mu_s, \mu_v)} \mathbb{E}_{(s, v) \sim \gamma, (s', v') \sim \gamma} [c(s, v)\!+\!\mathcal{L}(s, v, s', v')] \\
&=\!\mathop{min}\limits_{T \in \Pi(p^s, p^v)} \sum\limits_{i,i',j,j'} T_{ij}[\lambda c(s_i, v_i)\!+\!T_{i', j'}\mathcal{L}(s_i, v_j, s_{i'}, v_{j'})]
\end{align}
\end{small}
\noindent
where $\lambda$ is the weight hyper-parameter, $c(s_i, v_j)$ is the cost function that evaluates the node similarity between $s_i$ and $v_j$ using cosine distance, and $\mathcal{L}(s_i, v_j, s_{i'}, v_{j'}) = ||c_1(s_i, s_{i'}) - c_2(v_j, v_{j'})||$ is the cost function evaluating the similarity between two pairs of nodes $(s_i, s_{i'})$ and $(v_j, v_{j'})$.

We apply the Sinkhorn algorithm~\cite{cuturi2013sinkhorn, peyre2019computational} to obtain the optimal transport distance $\mathcal{D}(\mu, \nu)$, following \cite{chen2020graph, alvarez2018gromov}. Then, the calculated optimal transport distance is used as the cross-modal semantic coherence loss:  

\begin{equation}
\mathcal{L}_{cm} = \alpha \mathcal{D}(\mu, \nu) 
\end{equation}

\noindent
where $\alpha$ is the weight hyper-parameter. $\mathcal{L}_{cm}$ provides an explicit training objective to encourage semantic alignment of each sub-graph.

\vspace{-0.3cm}
\subsubsection{Cross-level Semantic Coherence}
Every video clip is composed of a sequence of segments, which follow a consistent theme intrinsically and narrate the event coherently. Motivated by this fact, we present the cross-level semantic coherence to promote the semantic coherence across temporal-level nodes and global-level nodes. By keeping the cross-level semantic coherence, we can greatly improve the graph representation's quality for story-level understanding.

Mutual information have been widely utilized for representation learning~(\eg variational autoencoder~\cite{kingma2013auto}, beta-VAE~\cite{higgins2016beta}), while we are the first to exploit mutual information to learn semantic coherence of the video. For temporal-level nodes $\{ t_i^{(n)}  \}_{i=1}^M$ and its corresponding global node $o_n$, we maximize the average mutual information between them as: $\frac{1}{M \times N} \sum_{n=1}^N  \sum_{i=1}^M \mathcal{I}(t_i^{(n)}; o_n)$. The mutual information maximization procedure can encourage segment-level reasoning and temporal-level reasoning to encode more underlying semantic information that is coherent in the video. To compute the mutual information, we use Noise-Contrastive Estimation~(NCE)~\cite{gutmann2012noise, gutmann2010noise, oord2018representation} to estimate it as follows:

\begin{equation}
\hat{\mathcal{I}}(t_i^{(n)}; o_n) := \mathbb{E}_{\mathbb{P}} [T_{\varphi}(t_i^{(n)}, o_n) - \mathbb{E}_{\tilde{\mathbb{P}}}[log\sum_{t_i^{(n)'}} e^{T_{\varphi}(t_i^{(n)'}, o_n)} ] ]
\end{equation}

\noindent
where $T_{\varphi}$ is a discriminator modeled by a neural network with parameters $\varphi$, $\mathbb{P}$ and $\tilde{\mathbb{P}}$ are the distribution of $t_i^{(n)}$. Then the cross-level semantic coherence can be formulated as:

\begin{equation}
\mathcal{L}_{cl} =  - \beta \frac{1}{M \times N} \sum_{n=1}^N \sum_{i=1}^M \hat{\mathcal{I}}(t_i^{(n)}; o_n)
\end{equation}

%\begin{equation}
%	<\theta^*, \varphi^*> = \mathop{argmax}\limits_{\theta, \varphi} \frac{1}{M \times N} \sum_{n=1}^N \sum_{i=1}^M \hat{\mathcal{I}}(t_i^{(n)}; o_n)
%\end{equation}

\noindent
where $\beta$ is weight hyper-parameter. We add this term to the loss function and optimize $\theta$ and $\varphi$ simultaneously. Thus, the total loss is given by:
\begin{equation}
\mathcal{L} = \mathcal{L}_{ent} + \mathcal{L}_{qe} +  \mathcal{L}_{cm} + \mathcal{L}_{cl}
\end{equation}

\noindent
where $\mathcal{L}_{ent}$ is the original cross-entropy loss.

\section{Experiments}

\subsection{Experimental Setup}

\noindent
\textbf{Dataset}\quad The VIOLIN dataset contains 15887 video clips collected from 4 popular TV shows and movie clips from YouTube channels covering thousands of movies. The average length of each video clip is 35.20s with 3 frames per second and each statement has 18 words on average. Each video clip is annotated with 3 pairs of positive/negative statements, resulting in 95322 $(V, S, H)$ triplets in total. It is divided into 76122, 9600, and 9600 triplets for training, validation, and testing, respectively. Model performance is evaluated via binary classification accuracy.

\noindent
\textbf{Implementation Details}\quad The dimension of the node embedding is set to 512. For the visual frames that are not paired with any subtitles, we assign it to the neighboring frame-subtitle pair. We set the halting threshold $\epsilon$ to 0.1, the maximum query number $N_{max}$ to 5, and the query efficiency $\tau$ to 0.05. In the training stage, the learning rate is $1e^{-4}$ and the batch size is 128.

\begin{table}[t]
	\resizebox{\linewidth}{!}{
		\begin{threeparttable}
			\begin{tabular}{ ccccccccccc}
				\toprule
				&\multicolumn{6}{c}{Method} & Vision  & Text  &Accuracy\\
				
%				\midrule
%				1 &\multicolumn{6}{c}{Random}          &-                 &-                       &50.00        \\
%				\midrule
%				2&\multicolumn{6}{c}{MTS Stmt}            &-                 &GloVe                &53.94        \\
%				3&\multicolumn{6}{c}{MTS Stmt}            &-                 &BERT                 &54.20         \\
%				\midrule
%				4&\multicolumn{6}{c}{MTS Stmt+Subtt} &-                 &GloVe                &60.10         \\
%				5&\multicolumn{6}{c}{MTS Stmt+Subtt} &-                 &BERT                 &66.05         \\
%				\midrule
%				6&\multicolumn{6}{c}{MTS Stmt+Vis}     &Img             &GloVe               &55.30         \\
%				7&\multicolumn{6}{c}{MTS Stmt+Vis}     &Img             &BERT                &59.26         \\
%				%				MTS Stmt+Vis     &C3D            &GloVe                &55.91         \\
%				8&\multicolumn{6}{c}{MTS Stmt+Vis}     &C3D             &BERT                &58.34         \\
%				%				MTS Stmt+Vis     &Det             &GloVe                &56.15        \\
%				9&\multicolumn{6}{c}{MTS Stmt+Vis}     &Det             &BERT                &59.45         \\
				
				\midrule    
				1&\multicolumn{6}{c}{MTS}     &Img             &GloVe                &60.33         \\
				2&\multicolumn{6}{c}{MTS}     &Img             &BERT                &67.60         \\
				%				MTS Stmt+Subtt+Vis     &C3D            &GloVe                &60.68         \\
				3&\multicolumn{6}{c}{MTS}     &C3D            &BERT                &67.23         \\
				%				MTS Stmt+Subtt+Vis     &Det             &GloVe                &61.31         \\
				4&\multicolumn{6}{c}{MTS}     &Det             &BERT                &67.84         \\
				
				\midrule
				5&\multicolumn{6}{c}{DIFFPOOL-Split}				&Img             &BERT                &59.46         \\
				6&\multicolumn{6}{c}{DIFFPOOL-Whole}		 &Img             &BERT                &56.43         \\
				
				\midrule
				7&\multicolumn{6}{c}{XML}				                  &Img             &BERT                &66.32         \\
				%17&\multicolumn{6}{c}{HERO 1-layer}				  &Img             &BERT                &68.23         \\
				%18&\multicolumn{6}{c}{HERO 2-layer}				 &Img             &BERT                &67.47         \\
				8&\multicolumn{6}{c}{HERO (pre-trained)}				                  &Img             &BERT                &68.59         \\
				
				\midrule
				9&\multicolumn{6}{c}{\textbf{Ours- AHGN + SCL}}    						   &Img             &BERT                &\textbf{71.38}         \\
				
				\bottomrule
			\end{tabular}
		\end{threeparttable}
	}
	\vspace{-0.2cm}
	\caption{Quantitative results on the VIOLIN dataset.}
	\label{t1}
	\vspace{-0.5cm}
\end{table} 

\noindent
\textbf{Baselines}\quad
%We compare our method with the following baselines: 
1-4)~MTS: these baselines are based on the multi-stream architecture~(please refer to Sec \ref{s2} for details).
 %which takes multiple streams of information as input (\eg statements, subtitles), maps the encoded features to the same feature space, computes the fusion features between pairs (statement-video pair, statement-subtitle pair) using bidirectional attention~\cite{lei2018tvqa, seo2016bidirectional, yu2018qanet}, and finally uses the fusion features to make the prediction. For visual feature, MTS also uses C3D features from 3-dimensional convolutional neural network~\cite{tran2015learning} and object detection features (Det) from Faster R-CNN~\cite{ren2015faster} trained on Visual Genome~\cite{krishna2017visual}.
We also compare our model with the state-of-the-art hierarchical graph representation learning method and hierarchical transformer-based models for video modeling: 5,6)~DIFFPOOL: a differentiable graph pooling module~\cite{ying2018hierarchical} that learns soft cluster assignment matrix for nodes in each layer. We try two versions: DIFFPOOL-whole that uniformly regards each video frame and subtitle word as the node, and DIFFPOOL-split that constructs a video frame graph and a subtitle word graph respectively. 7)~XML: Cross-modal Moment Localization (XML) modular network~\cite{lei2020tvr} is a recently proposed transformer-based method for TV show retrieval. 8)~HERO: a transformer-based framework~\cite{li2020hero} for video-and-language pre-training. It has two standard hierarchies with fixed structures for local and global context computation. %Note that HERO is first pre-trained on large-scale pre-training dataset using four pre-training tasks. (For the details of how we adapt these methods to VLI, please refer to the supplementary material.)

\subsection{Results}
We summarize the results in Table \ref{t1}, where our method significantly outperforms all baselines. Compared with the original baseline presented in \cite{liu2020violin}, our method surpasses it by $6.69\%$ relatively on accuracy. The comparison with DIFFPOOL indicates the effectiveness of our semantics-guided graph pooling scheme. The semantics-guided graph pooling can better control the complementary information fusion of subtitles and visual frames. Furthermore, XML models the relationships directly on the whole sequences of frames and subtitles, while we model the interaction in three hierarchies, leveraging the temporal alignment and complementary nature. The results show the efficiency of our hierarchical strategy. HERO first fuses every frame-subtitle pairs using a cross-modal transformer and then applies temporal transformer on the merged sequence of them. Our method differs as it represents the context at three different levels of granularity with adapative graph structure, which outperforms the pre-trained HERO by 2.79\%.

%However, HERO is also inherently flat and does not learn hierarchical semantics of the video. Our method shows better performance on modeling the global semantics in three hierarchies.

\begin{table}[t]
	\resizebox{\linewidth}{!}{
		\begin{threeparttable}
			\begin{tabular}{ lllllllcccc}
				\toprule
				&\multicolumn{6}{l}{Method} & Vision  & Text  &Accuracy\\
				
				\midrule    
				%1&\multicolumn{6}{l}{MTS Stmt+Subtt+Vis}     &Img             &GloVe                &60.33         \\
				1&\multicolumn{6}{l}{MTS}     &Img             &BERT                &67.60         \\
				%				MTS Stmt+Subtt+Vis     &C3D            &GloVe                &60.68         \\
				%3&\multicolumn{6}{l}{MTS Stmt+Subtt+Vis}     &C3D            &BERT                &67.23         \\
				%				MTS Stmt+Subtt+Vis     &Det             &GloVe                &61.31         \\
				%4&\multicolumn{6}{l}{MTS Stmt+Subtt+Vis}     &Det             &BERT                &67.84         \\
				
				\midrule
				2&\multicolumn{6}{l}{AHGN}    				    &Img             &BERT                &69.76         \\
				3&\multicolumn{6}{l}{AHGN + $\mathcal{L}_{cm}$}    				    &Img             &BERT                &70.19         \\
				4&\multicolumn{6}{l}{AHGN + $\mathcal{L}_{cl}$}    				    &Img             &BERT                &70.47        \\
				5&\multicolumn{6}{l}{AHGN + $\mathcal{L}_{cm}$ + $\mathcal{L}_{cl}$ }    				    &Img             &BERT                &\textbf{71.38}         \\
				\bottomrule
			\end{tabular}
		\end{threeparttable}
	}
	\vspace{-0.2cm}
	\caption{Main ablation results.}
	\label{t2}
	\vspace{-0.3cm}
\end{table}

\begin{figure}[!t]
	\centering
	\includegraphics[width=\linewidth]{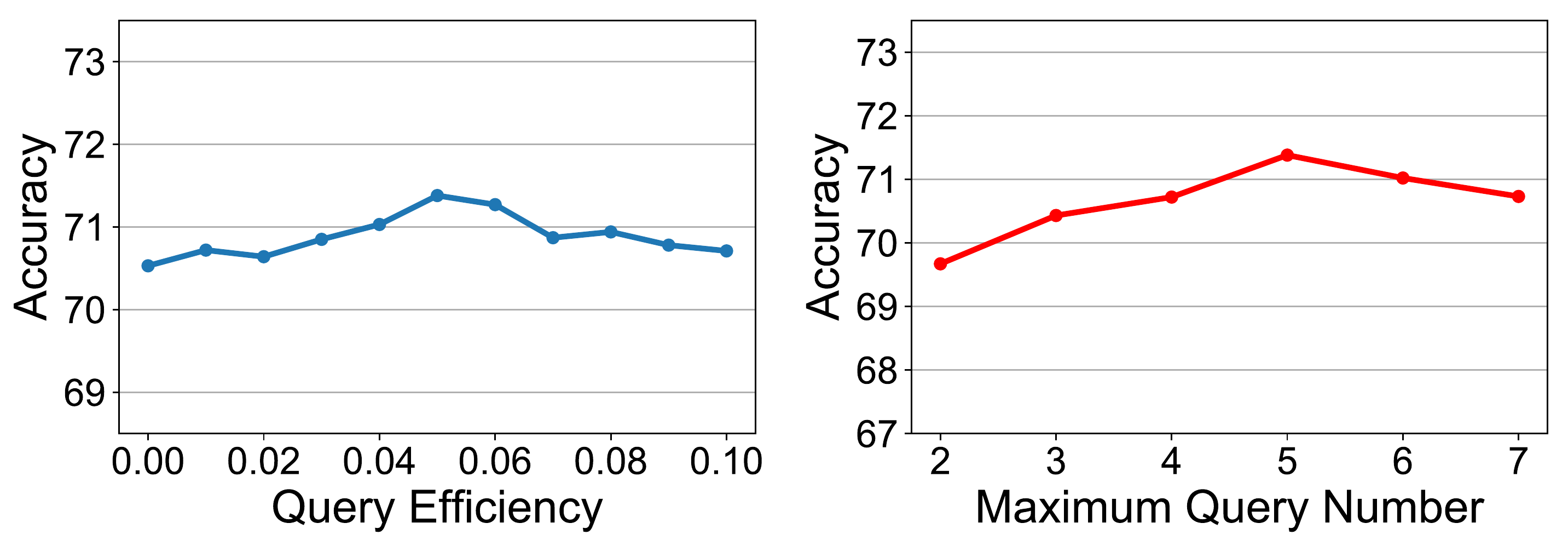}
	\vspace{-0.5cm}
	\caption{Ablation across with respect to $\tau$ and $N_{max}$.}
	%We provide the results when we use different number of sub-policies.
	\label{ablation}
	\vspace{-0.5cm}
\end{figure}

\subsection{In-Depth Analysis}

\noindent
\textbf{Effectiveness of Individual Components}\quad We conduct an ablation study to illustrate the effectiveness of each component in Table \ref{t2}. Comparing MTS and AHGN~(Row 1 vs Row 2), AHGN significantly contributes 2.16\% to the improvement on accuracy. The results of Row 3 and Row 4 validate the superiority of the cross-modal and cross-level semantic coherence, respectively. Meanwhile, the results indicate that the introduced two losses can promote the cross-modal and cross-level semantic coherence of the AHGN in a mutually rewarding way. Finally, the semantic coherence learning~(Row 5) takes up 2.32\% of the relative gain on accuracy.

\begin{figure*}
	\begin{center}
		\includegraphics[width=\textwidth]{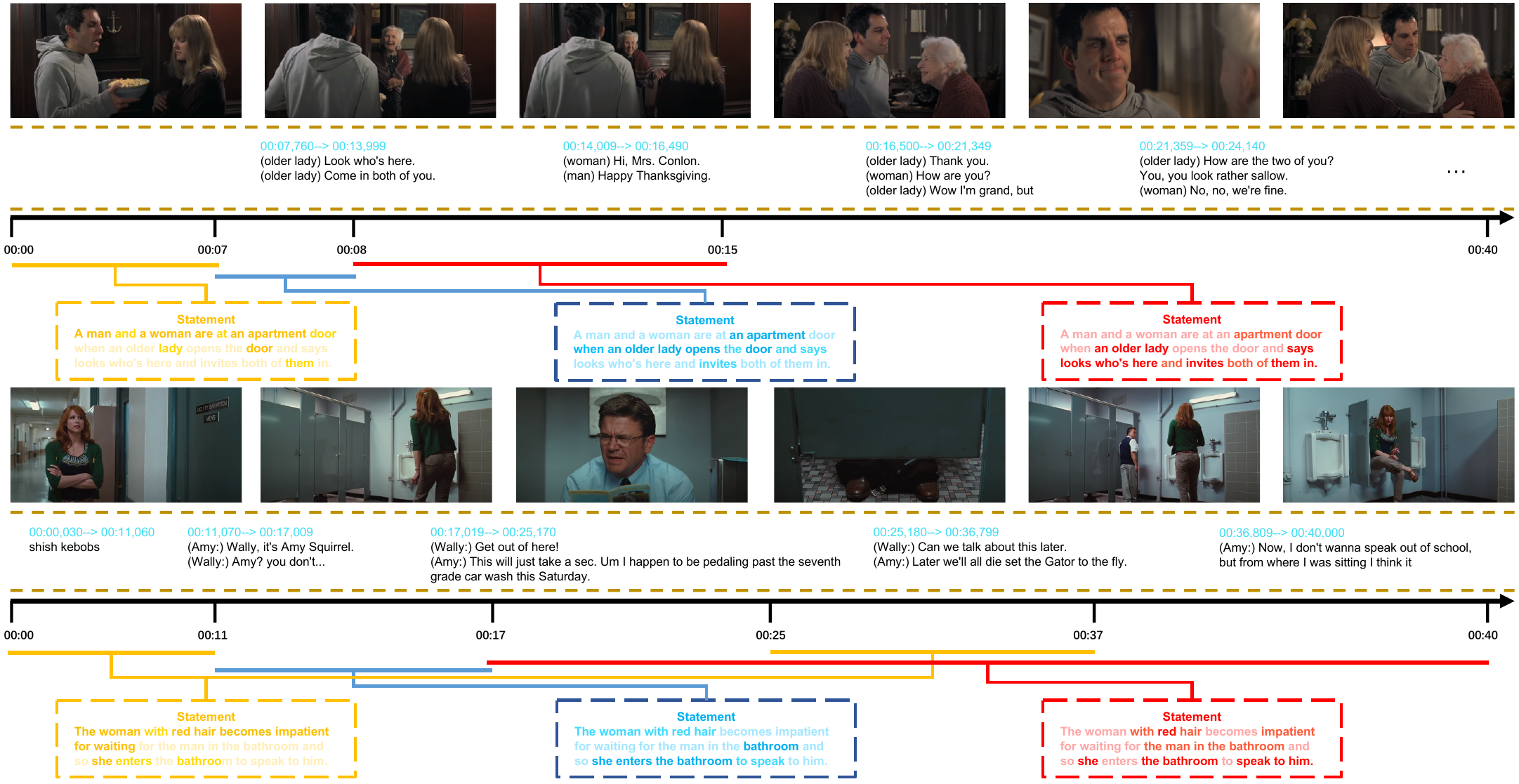}
	\end{center}
	\vspace{-0.3cm}
	\caption{Qualitative Examples. The attention weights of different semantic queries are illustrated by the depth of color, and their corresponding video moments with high attention weights during semantics-guided graph pooling are also shown.}
	\label{qualitative}
	\vspace{-0.25cm}
\end{figure*}

\begin{figure}[!t]
	\centering
	\footnotesize
	\includegraphics[width=\linewidth]{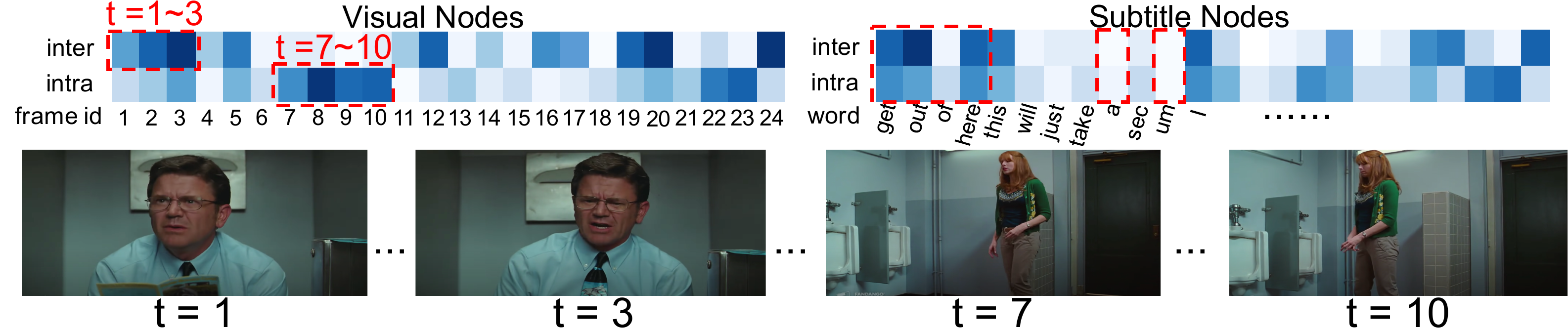}
	\vspace{-0.5cm}
	\caption{Visualization of the GER and GRA.}
	\label{v1}
	\vspace{-0.6cm}
\end{figure}

\noindent
\textbf{Analysis on AHGN}\quad We further perform in-depth analysis for the adaptive hierarchical graph network. We investigate the contribution of the proposed three graph operations and the adaptive graph structure. We start with the backbone model, which keeps the adaptive hierarchical structure but removes the proposed graph operations. For the ablation model with fixed graph structure, we set the number of temporal-level sub-graphs as a hyper-parameter. Specifically, we extract fixed number of semantic queries from the statement and use them to construct temporal-level sub-graphs respectively. We test different number of temporal-level sub-graphs and report the best performance when the number of temporal-level sub-graph is 3. Note that the detailed results are provided in the supplementary material. For the ablation model without temporal-level reasoning, we directly perform attentive pooling using semantic queries on the nodes after segment-level reasoning.

\begin{table}[t]
	\resizebox{\linewidth}{!}{
		\begin{threeparttable}
			\begin{tabular}{ lllllll|ccc|cc|c}
				\toprule
				&\multicolumn{6}{l|}{\multirow{2}*{Method}} &\multicolumn{3}{c|}{Graph Operation} &\multicolumn{2}{c|}{Graph Structure}   &\multirow{2}*{Accuracy} \\
				&\multicolumn{6}{l|}{} & GER  & GRA  &Temp &fixed &adaptive &\\
				
				\hline    
				1&\multicolumn{6}{l|}{Backbone (fixed)}     &       &         &    &\Checkmark  &    &64.65         \\
				2&\multicolumn{6}{l|}{Backbone}     &       &         &    & &\Checkmark    &65.32         \\
				
				\hline
				3&\multicolumn{6}{l|}{\quad + GER}    		    &\Checkmark        &     &    & &\Checkmark    &67.07         \\
				4&\multicolumn{6}{l|}{\quad + GRA}    		     &         &\Checkmark      &    & &\Checkmark   &66.74        \\
				5&\multicolumn{6}{l|}{\quad + Temporal}    	  &    &     &\Checkmark    &     &\Checkmark &66.93  \\
				
				\hline
				6&\multicolumn{6}{l|}{\quad + GER + GRA}    	&\Checkmark        &\Checkmark     & &  &\Checkmark           &68.03         \\
				7&\multicolumn{6}{l|}{\quad + GER + Temp}     &\Checkmark                & & \Checkmark    &  &\Checkmark       &68.64        \\
				8&\multicolumn{6}{l|}{\quad + GRA + Temp}    &             &\Checkmark      &\Checkmark   &  &\Checkmark    &68.15         \\
				
				\hline
				9&\multicolumn{6}{l|}{AHGN}    	 &\Checkmark  &\Checkmark  &\Checkmark & &\Checkmark
				&69.76         \\ 
				10&\multicolumn{6}{l|}{AHGN (fixed)}    	 &\Checkmark  &\Checkmark  &\Checkmark &\Checkmark &
				&68.91         \\      
				\bottomrule
			\end{tabular}
		\end{threeparttable}
	}
	\vspace{-0.2cm}
	\caption{We conduct comparison by varying the individual components of the AHGN.}
	\label{t3}
	\vspace{-0.6cm}
\end{table}

%We start with the AHGN and replace the gated inter-modal message passing~(GER), the gated intra-modal message passing~(GRA), the context gate in inter- and intra-modal message passing, and the temporal-level reasoning, respectively. We also validate the superiority of the adaptive graph structure by using fixed graph structure. For the ablation model without the context gate, as \cite{hamilton2017inductive}, we concatenate the node's current representation with the aggregated message and feed the concatenated vector through a fully connected layer without nonlinear activation function, which transforms the representation to the original dimension. %For the ablation model without SGP, we perform temporal-level reasoning over the whole sequences of the subtitle nodes and visual nodes respectively, then get the global subtitle node and visual node using mean pooling, and finally concatenate them to get the corresponding global graph representation. For the ablation model without temporal-level reasoning, we directly perform attentive pooling using semantic queries on the nodes after segment-level reasoning. For the ablation model using fixed graph structure, we set the number of temporal-level sub-graphs as a hyper-parameter. We test different number of temporal-level sub-graph and report the best performance when the number of temporal-level sub-graph is 3. (More detailed results are provided in the supplementary material.)

Table \ref{t3} summarizes the results, which indicate the following. First, the adaptive graph structure is more effective than the fixed graph structure, which enables our AHGN to dynamically adjust the graph structure according to the statement. Second, the inter- and intra- modality reasoning at segment-level significantly improve the performance by modeling the inherent alignment and complementary nature between visual frames and subtitles. Third, the temporal-level reasoning is a crucial step to achieve in-depth understanding of the video.

%Secondly, the accuracy of the ablation model without the context gate drops 0.99 points, which implies the importance of the context gate to control the information flow between subtitle nodes and visual nodes. Thirdly, the semantics-guided graph pooling further improves the performance and is more efficient than directly reasoning over the whole sequences. Thirdly, the temporal-level reasoning is a crucial step to achieve in-depth understanding of the video. Finally, the adaptive graph structure enables our AHGN to dynamically adjust the graph structure according to the statement, which further improves the performance.

\noindent
\textbf{Ablation of the Adaptive Graph Construction}\quad We explore the impact of the query efficiency hyper-parameter $\tau$ and the maximum query number $N_{max}$ for the adaptive graph construction~(AGC). The higher $\tau$ means that the tolerance for more queries is lower. As illustrated in Figure \ref{ablation}, the performance keeps increasing when the $\tau$ is increased from 0.02 to 0.05. When we continue to increase the $\tau$, the performance decreases because too large $\tau$ limits the number of semantic queries to one or two. Further, the maximum query number $N_{max}$ of 5 and 6 are enough and can generally provide good performance while higher $N_{max}$ does harm to the performance and reduces the efficiency. %To guarantee the performance and reduce the computational complexity, we set the query efficiency $\tau$ to 0.05 and set the maximum query number $N_{max}$ to 5. 

%Also, we evaluate the ablation model that uses a fixed number of queries and find that our ASG achieves better performance. Due to the space limit, the detailed results are provided in the supplementary material.

\subsection{Qualitative Analysis}
For a more intuitive view of how our model works for the VLI task, we visualize two qualitative examples in Figure~\ref{qualitative}. %where the attention weights of different semantic queries and the most concerned video moments during semantics-guided graph pooling are illustrated. 
The attention weights of semantic queries reflect the semantic parts which their corresponding temporal-level sub-graphs focused on. As shown in Figure \ref{qualitative}, different temporal-level sub-graphs focus on different semantic phrases of the statement and pay attention to the video segments that are most related to their semantics.

\begin{figure}[!t]
	\centering
	\footnotesize
	\includegraphics[width=0.9\linewidth]{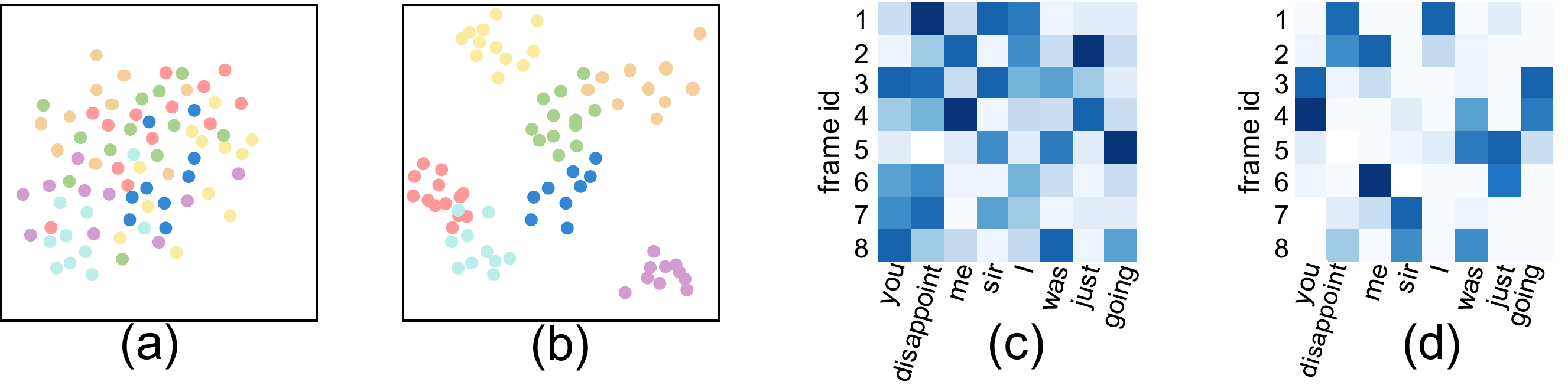}
	\vspace{-0.1cm}
	\caption{(a)(b): t-SNE visualization of temporal-level nodes without/with $\mathcal{L}_{cl}$. (c)(d): alignment matrixes without/with $\mathcal{L}_{cm}$.}
	\label{v2}
	\vspace{-0.5cm}
\end{figure}

\noindent
\textbf{Visualization}\quad Fig \ref{v1} visualizes the inter- and intra- modality gate values 
%of a paired visual nodes and subtitle nodes
~(top row). Some key frames are in the bottom row. The $t=1\sim3$ frames show that a man talks to someone angrily, and the inter-modality gates are activated to combine linguistic context from subtitles. This helps the model to understand what he says and why he is angry. For the $t=7\sim10$ frames, the intra-modality gates are well activated to combine temporal information from local visual context. Thereby, the model can infer that the woman is coming in instead of standing still. Also, gate values of words ``get'' and ``here'' are both high. It indicates that GER and GRA cooperate with each other to infer the complete semantics and visual context. Some words like ``a" and ``um'' receive low gate values. Fig \ref{v2} provides visualization on temporal-level node feature space and learned cross-modality alignment matrices between visual and subtitle nodes. 
We observe that \textbf{with $\mathcal{L}_{cl}$} (Fig \ref{v2}.b), the nodes from the same temporal-level sub-graph~(nodes of the same color) tend to be more tightly related, compared with the features trained \textbf{without $\mathcal{L}_{cl}$} (Fig \ref{v2}.a). \textbf{Without $\mathcal{L}_{cm}$} (Fig \ref{v2}.c), the learned alignment matrix is much denser and noisier than the alignment matrix learned \textbf{with $\mathcal{L}_{cm}$} (Fig \ref{v2}.d).

\vspace{-0.1cm}
\section{Conclusions}
In this paper, we introduce an adaptive hierarchical graph reasoning with semantic coherence approach for Video-and-Language Inference. Our  adaptive hierarchical graph network performs in-depth reasoning over video frames and subtitles in three hierarchies, where the graph reasoning structure is adaptively determined by the semantic structures of the statement. Further, we present a semantic coherence learning algorithm to encourage the cross-modal and cross-level semantic coherence of the adaptive hierarchical graph network. The experimental results show that our method outperforms the baselines by a large margin.

%\vspace{-0.3cm}
\section*{Acknowledgment}
This work has been supported in part by NSFC (No. 61625107), Zhejiang NSF (LR21F020004), National Key Research and Development Program of China (2018AAA0101900), Alibaba-Zhejiang University Joint Research Institute of Frontier Technologies, Zhejiang University iFLYTEK Joint Research Center, Chinese Knowledge Center of Engineering Science and Technology (CKCEST).

\clearpage
{\small

\bibliographystyle{ieee_fullname}
\bibliography{egbib}

\begin{thebibliography}{10}\itemsep=-1pt

\bibitem{abu2016youtube}
Sami Abu-El-Haija, Nisarg Kothari, Joonseok Lee, Paul Natsev, George Toderici,
  Balakrishnan Varadarajan, and Sudheendra Vijayanarasimhan.
\newblock Youtube-8m: A large-scale video classification benchmark.
\newblock {\em arXiv preprint arXiv:1609.08675}.

\bibitem{alvarez2018gromov}
David Alvarez-Melis and Tommi~S Jaakkola.
\newblock Gromov-wasserstein alignment of word embedding spaces.
\newblock {\em arXiv preprint arXiv:1809.00013}, 2018.

\bibitem{anne2017localizing}
Lisa Anne~Hendricks, Oliver Wang, Eli Shechtman, Josef Sivic, Trevor Darrell,
  and Bryan Russell.
\newblock Localizing moments in video with natural language.
\newblock In {\em Proceedings of the IEEE international conference on computer
  vision}, pages 5803--5812, 2017.

\bibitem{antol2015vqa}
Stanislaw Antol, Aishwarya Agrawal, Jiasen Lu, Margaret Mitchell, Dhruv Batra,
  C~Lawrence Zitnick, and Devi Parikh.
\newblock Vqa: Visual question answering.
\newblock In {\em Proceedings of the IEEE international conference on computer
  vision}, pages 2425--2433, 2015.

\bibitem{caba2015activitynet}
Fabian Caba~Heilbron, Victor Escorcia, Bernard Ghanem, and Juan Carlos~Niebles.
\newblock Activitynet: A large-scale video benchmark for human activity
  understanding.
\newblock In {\em Proceedings of the ieee conference on computer vision and
  pattern recognition}, pages 961--970, 2015.

\bibitem{chen2018temporally}
Jingyuan Chen, Xinpeng Chen, Lin Ma, Zequn Jie, and Tat-Seng Chua.
\newblock Temporally grounding natural sentence in video.
\newblock In {\em Proceedings of the 2018 conference on empirical methods in
  natural language processing}, pages 162--171, 2018.

\bibitem{chen2020graph}
Liqun Chen, Zhe Gan, Yu Cheng, Linjie Li, Lawrence Carin, and Liu Jingjing.
\newblock Graph optimal transport for cross-domain alignment.
\newblock {\em arXiv preprint arXiv:2006.14744}, 2020.

\bibitem{cheng2014temporal}
Yu Cheng, Quanfu Fan, Sharath Pankanti, and Alok Choudhary.
\newblock Temporal sequence modeling for video event detection.
\newblock In {\em Proceedings of the IEEE Conference on Computer Vision and
  Pattern Recognition}, pages 2227--2234, 2014.

\bibitem{chowdhury2019gromov}
Samir Chowdhury and Facundo M{\'e}moli.
\newblock The gromov--wasserstein distance between networks and stable network
  invariants.
\newblock {\em Information and Inference: A Journal of the IMA}, 8(4):757--787,
  2019.

\bibitem{cuturi2013sinkhorn}
Marco Cuturi.
\newblock Sinkhorn distances: Lightspeed computation of optimal transport.
\newblock In {\em Advances in neural information processing systems}, pages
  2292--2300, 2013.

\bibitem{deng2009imagenet}
Jia Deng, Wei Dong, Richard Socher, Li-Jia Li, Kai Li, and Li Fei-Fei.
\newblock Imagenet: A large-scale hierarchical image database.
\newblock In {\em 2009 IEEE conference on computer vision and pattern
  recognition}, pages 248--255. Ieee, 2009.

\bibitem{devlin2018bert}
Jacob Devlin, Ming-Wei Chang, Kenton Lee, and Kristina Toutanova.
\newblock Bert: Pre-training of deep bidirectional transformers for language
  understanding.
\newblock {\em arXiv preprint arXiv:1810.04805}, 2018.

\bibitem{gan2017stylenet}
Chuang Gan, Zhe Gan, Xiaodong He, Jianfeng Gao, and Li Deng.
\newblock Stylenet: Generating attractive visual captions with styles.
\newblock In {\em Proceedings of the IEEE Conference on Computer Vision and
  Pattern Recognition}, pages 3137--3146, 2017.

\bibitem{gan2017semantic}
Zhe Gan, Chuang Gan, Xiaodong He, Yunchen Pu, Kenneth Tran, Jianfeng Gao,
  Lawrence Carin, and Li Deng.
\newblock Semantic compositional networks for visual captioning.
\newblock In {\em Proceedings of the IEEE conference on computer vision and
  pattern recognition}, pages 5630--5639, 2017.

\bibitem{gao2017tall}
Jiyang Gao, Chen Sun, Zhenheng Yang, and Ram Nevatia.
\newblock Tall: Temporal activity localization via language query.
\newblock In {\em Proceedings of the IEEE international conference on computer
  vision}, pages 5267--5275, 2017.

\bibitem{guadarrama2013youtube2text}
Sergio Guadarrama, Niveda Krishnamoorthy, Girish Malkarnenkar, Subhashini
  Venugopalan, Raymond Mooney, Trevor Darrell, and Kate Saenko.
\newblock Youtube2text: Recognizing and describing arbitrary activities using
  semantic hierarchies and zero-shot recognition.
\newblock In {\em Proceedings of the IEEE international conference on computer
  vision}, pages 2712--2719, 2013.

\bibitem{guo2021semi-supervised}
Jiannan Guo, Haochen Shi, Yangyang Kang, Kun Kuang, Siliang Tang, Zhuoren
  Jiang, Changlong Sun, Fei Wu, and Yueting Zhuang.
\newblock Semi-supervised active learning for semi-supervised models: Exploit
  adversarial examples with graph-based virtual labels.
\newblock In {\em Proceedings of the IEEE/CVF International Conference on
  Computer Vision}, 2021.

\bibitem{gutmann2010noise}
Michael Gutmann and Aapo Hyv{\"a}rinen.
\newblock Noise-contrastive estimation: A new estimation principle for
  unnormalized statistical models.
\newblock In {\em Proceedings of the Thirteenth International Conference on
  Artificial Intelligence and Statistics}, pages 297--304, 2010.

\bibitem{gutmann2012noise}
Michael~U Gutmann and Aapo Hyv{\"a}rinen.
\newblock Noise-contrastive estimation of unnormalized statistical models, with
  applications to natural image statistics.
\newblock {\em The journal of machine learning research}, 13(1):307--361, 2012.

\bibitem{he2016deep}
Kaiming He, Xiangyu Zhang, Shaoqing Ren, and Jian Sun.
\newblock Deep residual learning for image recognition.
\newblock In {\em Proceedings of the IEEE conference on computer vision and
  pattern recognition}, pages 770--778, 2016.

\bibitem{heo2019constructing}
Yu-Jung Heo, Kyoung-Woon On, Seongho Choi, Jaeseo Lim, Jinah Kim, Jeh-Kwang
  Ryu, Byung-Chull Bae, and Byoung-Tak Zhang.
\newblock Constructing hierarchical q\&a datasets for video story
  understanding.
\newblock {\em arXiv preprint arXiv:1904.00623}, 2019.

\bibitem{higgins2016beta}
Irina Higgins, Loic Matthey, Arka Pal, Christopher Burgess, Xavier Glorot,
  Matthew Botvinick, Shakir Mohamed, and Alexander Lerchner.
\newblock beta-vae: Learning basic visual concepts with a constrained
  variational framework.
\newblock 2016.

\bibitem{jang2017tgif}
Yunseok Jang, Yale Song, Youngjae Yu, Youngjin Kim, and Gunhee Kim.
\newblock Tgif-qa: Toward spatio-temporal reasoning in visual question
  answering.
\newblock In {\em Proceedings of the IEEE Conference on Computer Vision and
  Pattern Recognition}, pages 2758--2766, 2017.

\bibitem{kay2017kinetics}
Will Kay, Joao Carreira, Karen Simonyan, Brian Zhang, Chloe Hillier, Sudheendra
  Vijayanarasimhan, Fabio Viola, Tim Green, Trevor Back, Paul Natsev, et~al.
\newblock The kinetics human action video dataset.
\newblock {\em arXiv preprint arXiv:1705.06950}, 2017.

\bibitem{kim2017deepstory}
Kyung-Min Kim, Min-Oh Heo, Seong-Ho Choi, and Byoung-Tak Zhang.
\newblock Deepstory: Video story qa by deep embedded memory networks.
\newblock {\em arXiv preprint arXiv:1707.00836}, 2017.

\bibitem{kingma2013auto}
Diederik~P Kingma and Max Welling.
\newblock Auto-encoding variational bayes.
\newblock {\em arXiv preprint arXiv:1312.6114}, 2013.

\bibitem{krishna2017dense}
Ranjay Krishna, Kenji Hata, Frederic Ren, Li Fei-Fei, and Juan Carlos~Niebles.
\newblock Dense-captioning events in videos.
\newblock In {\em Proceedings of the IEEE international conference on computer
  vision}, pages 706--715, 2017.

\bibitem{lei2018tvqa}
Jie Lei, Licheng Yu, Mohit Bansal, and Tamara~L Berg.
\newblock Tvqa: Localized, compositional video question answering.
\newblock {\em arXiv preprint arXiv:1809.01696}, 2018.

\bibitem{lei2020tvr}
Jie Lei, Licheng Yu, Tamara~L Berg, and Mohit Bansal.
\newblock Tvr: A large-scale dataset for video-subtitle moment retrieval.
\newblock {\em arXiv preprint arXiv:2001.09099}, 2020.

\bibitem{li2019walking}
Juncheng Li, Siliang Tang, Fei Wu, and Yueting Zhuang.
\newblock Walking with mind: Mental imagery enhanced embodied qa.
\newblock In {\em Proceedings of the 27th ACM International Conference on
  Multimedia}, pages 1211--1219, 2019.

\bibitem{li2020unsupervised}
Juncheng Li, Xin Wang, Siliang Tang, Haizhou Shi, Fei Wu, Yueting Zhuang, and
  William~Yang Wang.
\newblock Unsupervised reinforcement learning of transferable meta-skills for
  embodied navigation.
\newblock In {\em Proceedings of the IEEE/CVF Conference on Computer Vision and
  Pattern Recognition}, pages 12123--12132, 2020.

\bibitem{li2020hero}
Linjie Li, Yen-Chun Chen, Yu Cheng, Zhe Gan, Licheng Yu, and Jingjing Liu.
\newblock Hero: Hierarchical encoder for video+ language omni-representation
  pre-training.
\newblock {\em arXiv preprint arXiv:2005.00200}, 2020.

\bibitem{liu2020violin}
Jingzhou Liu, Wenhu Chen, Yu Cheng, Zhe Gan, Licheng Yu, Yiming Yang, and
  Jingjing Liu.
\newblock Violin: A large-scale dataset for video-and-language inference.
\newblock In {\em Proceedings of the IEEE/CVF Conference on Computer Vision and
  Pattern Recognition}, pages 10900--10910, 2020.

\bibitem{liu2020semantic}
Yanbin Liu, Linchao Zhu, Makoto Yamada, and Yi Yang.
\newblock Semantic correspondence as an optimal transport problem.
\newblock In {\em Proceedings of the IEEE/CVF Conference on Computer Vision and
  Pattern Recognition}, pages 4463--4472, 2020.

\bibitem{luise2018differential}
Giulia Luise, Alessandro Rudi, Massimiliano Pontil, and Carlo Ciliberto.
\newblock Differential properties of sinkhorn approximation for learning with
  wasserstein distance.
\newblock In {\em Advances in Neural Information Processing Systems}, pages
  5859--5870, 2018.

\bibitem{mun2020local}
Jonghwan Mun, Minsu Cho, and Bohyung Han.
\newblock Local-global video-text interactions for temporal grounding.
\newblock In {\em Proceedings of the IEEE/CVF Conference on Computer Vision and
  Pattern Recognition}, pages 10810--10819, 2020.

\bibitem{mun2017marioqa}
Jonghwan Mun, Paul Hongsuck~Seo, Ilchae Jung, and Bohyung Han.
\newblock Marioqa: Answering questions by watching gameplay videos.
\newblock In {\em Proceedings of the IEEE International Conference on Computer
  Vision}, pages 2867--2875, 2017.

\bibitem{oord2018representation}
Aaron van~den Oord, Yazhe Li, and Oriol Vinyals.
\newblock Representation learning with contrastive predictive coding.
\newblock {\em arXiv preprint arXiv:1807.03748}, 2018.

\bibitem{peyre2019computational}
Gabriel Peyr{\'e}, Marco Cuturi, et~al.
\newblock Computational optimal transport: With applications to data science.
\newblock {\em Foundations and Trends{\textregistered} in Machine Learning},
  11(5-6):355--607, 2019.

\bibitem{peyre2016gromov}
Gabriel Peyr{\'e}, Marco Cuturi, and Justin Solomon.
\newblock Gromov-wasserstein averaging of kernel and distance matrices.
\newblock In {\em International Conference on Machine Learning}, pages
  2664--2672, 2016.

\bibitem{pu2016adaptive}
Yunchen Pu, Martin~Renqiang Min, Zhe Gan, and Lawrence Carin.
\newblock Adaptive feature abstraction for translating video to text.
\newblock {\em arXiv preprint arXiv:1611.07837}, 2016.

\bibitem{seo2016bidirectional}
Minjoon Seo, Aniruddha Kembhavi, Ali Farhadi, and Hannaneh Hajishirzi.
\newblock Bidirectional attention flow for machine comprehension.
\newblock {\em arXiv preprint arXiv:1611.01603}, 2016.

\bibitem{kai2021ask}
Kai Shen, Lingfei Wu, Siliang Tang, Fangli Xu, Zhu Zhang, Yu Qiang, and Yueting
  Zhuang.
\newblock Ask question with double hints: Visual question generation with
  answer-awareness and region-reference, 2021.

\bibitem{ijcai2020-131}
Kai Shen, Lingfei Wu, Fangli Xu, Siliang Tang, Jun Xiao, and Yueting Zhuang.
\newblock Hierarchical attention based spatial-temporal graph-to-sequence
  learning for grounded video description.
\newblock In Christian Bessiere, editor, {\em Proceedings of the Twenty-Ninth
  International Joint Conference on Artificial Intelligence, {IJCAI-20}}, pages
  941--947. International Joint Conferences on Artificial Intelligence
  Organization, 7 2020.
\newblock Main track.

\bibitem{su2015optimal}
Zhengyu Su, Yalin Wang, Rui Shi, Wei Zeng, Jian Sun, Feng Luo, and Xianfeng Gu.
\newblock Optimal mass transport for shape matching and comparison.
\newblock {\em IEEE transactions on pattern analysis and machine intelligence},
  37(11):2246--2259, 2015.

\bibitem{suhr2018corpus}
Alane Suhr, Stephanie Zhou, Ally Zhang, Iris Zhang, Huajun Bai, and Yoav Artzi.
\newblock A corpus for reasoning about natural language grounded in
  photographs.
\newblock {\em arXiv preprint arXiv:1811.00491}, 2018.

\bibitem{tapaswi2016movieqa}
Makarand Tapaswi, Yukun Zhu, Rainer Stiefelhagen, Antonio Torralba, Raquel
  Urtasun, and Sanja Fidler.
\newblock Movieqa: Understanding stories in movies through question-answering.
\newblock In {\em Proceedings of the IEEE conference on computer vision and
  pattern recognition}, pages 4631--4640, 2016.

\bibitem{venugopalan2015sequence}
Subhashini Venugopalan, Marcus Rohrbach, Jeffrey Donahue, Raymond Mooney,
  Trevor Darrell, and Kate Saenko.
\newblock Sequence to sequence-video to text.
\newblock In {\em Proceedings of the IEEE international conference on computer
  vision}, pages 4534--4542, 2015.

\bibitem{wang2016walk}
Jing Wang, Yu Cheng, and Rogerio~Schmidt Feris.
\newblock Walk and learn: Facial attribute representation learning from
  egocentric video and contextual data.
\newblock In {\em Proceedings of the IEEE conference on computer vision and
  pattern recognition}, pages 2295--2304, 2016.

\bibitem{wang2019vatex}
Xin Wang, Jiawei Wu, Junkun Chen, Lei Li, Yuan-Fang Wang, and William~Yang
  Wang.
\newblock Vatex: A large-scale, high-quality multilingual dataset for
  video-and-language research.
\newblock In {\em Proceedings of the IEEE International Conference on Computer
  Vision}, pages 4581--4591, 2019.

\bibitem{xie2019visual}
Ning Xie, Farley Lai, Derek Doran, and Asim Kadav.
\newblock Visual entailment: A novel task for fine-grained image understanding.
\newblock {\em arXiv preprint arXiv:1901.06706}, 2019.

\bibitem{xu2016msr}
Jun Xu, Tao Mei, Ting Yao, and Yong Rui.
\newblock Msr-vtt: A large video description dataset for bridging video and
  language.
\newblock In {\em Proceedings of the IEEE conference on computer vision and
  pattern recognition}, pages 5288--5296, 2016.

\bibitem{ying2018hierarchical}
Zhitao Ying, Jiaxuan You, Christopher Morris, Xiang Ren, Will Hamilton, and
  Jure Leskovec.
\newblock Hierarchical graph representation learning with differentiable
  pooling.
\newblock In {\em Advances in neural information processing systems}, pages
  4800--4810, 2018.

\bibitem{young2014image}
Peter Young, Alice Lai, Micah Hodosh, and Julia Hockenmaier.
\newblock From image descriptions to visual denotations: New similarity metrics
  for semantic inference over event descriptions.
\newblock {\em Transactions of the Association for Computational Linguistics},
  2:67--78, 2014.

\bibitem{yu2018qanet}
Adams~Wei Yu, David Dohan, Minh-Thang Luong, Rui Zhao, Kai Chen, Mohammad
  Norouzi, and Quoc~V Le.
\newblock Qanet: Combining local convolution with global self-attention for
  reading comprehension.
\newblock {\em arXiv preprint arXiv:1804.09541}, 2018.

\bibitem{zhang2020devlbert}
Shengyu Zhang, Tan Jiang, Tan Wang, Kun Kuang, Zhou Zhao, Jianke Zhu, Jin Yu,
  Hongxia Yang, and Fei Wu.
\newblock Devlbert: Learning deconfounded visio-linguistic representations.
\newblock In {\em Proceedings of the 28th ACM International Conference on
  Multimedia}, pages 4373--4382, 2020.

\bibitem{zhang2020poet}
Shengyu Zhang, Ziqi Tan, Jin Yu, Zhou Zhao, Kun Kuang, Jie Liu, Jingren Zhou,
  Hongxia Yang, and Fei Wu.
\newblock Poet: Product-oriented video captioner for e-commerce.
\newblock In {\em Proceedings of the 28th ACM International Conference on
  Multimedia}, pages 1292--1301, 2020.

\bibitem{zhang2020comprehensive}
Shengyu Zhang, Ziqi Tan, Zhou Zhao, Jin Yu, Kun Kuang, Tan Jiang, Jingren Zhou,
  Hongxia Yang, and Fei Wu.
\newblock Comprehensive information integration modeling framework for video
  titling.
\newblock In {\em Proceedings of the 26th ACM SIGKDD International Conference
  on Knowledge Discovery \& Data Mining}, pages 2744--2754, 2020.

\bibitem{zhang2021consensus}
Wenqiao Zhang, Haochen Shi, Siliang Tang, Jun Xiao, Qiang Yu, and Yueting
  Zhuang.
\newblock Consensus graph representation learning for better grounded image
  captioning.
\newblock In {\em Proc 35 AAAI Conf on Artificial Intelligence}, 2021.

\bibitem{zhang2019frame}
Wenqiao Zhang, Siliang Tang, Yanpeng Cao, Shiliang Pu, Fei Wu, and Yueting
  Zhuang.
\newblock Frame augmented alternating attention network for video question
  answering.
\newblock {\em IEEE Transactions on Multimedia}, 22(4):1032--1041, 2019.

\bibitem{zhang2020relational}
Wenqiao Zhang, Xin~Eric Wang, Siliang Tang, Haizhou Shi, Haochen Shi, Jun Xiao,
  Yueting Zhuang, and William~Yang Wang.
\newblock Relational graph learning for grounded video description generation.
\newblock In {\em Proceedings of the 28th ACM International Conference on
  Multimedia}, pages 3807--3828, 2020.

\bibitem{zhu2017uncovering}
Linchao Zhu, Zhongwen Xu, Yi Yang, and Alexander~G Hauptmann.
\newblock Uncovering the temporal context for video question answering.
\newblock {\em International Journal of Computer Vision}, 124(3):409--421,
  2017.

\bibitem{zhu2020actbert}
Linchao Zhu and Yi Yang.
\newblock Actbert: Learning global-local video-text representations.
\newblock In {\em Proceedings of the IEEE/CVF conference on computer vision and
  pattern recognition}, pages 8746--8755, 2020.

\end{thebibliography}
}

\end{document}